\documentclass[11pt,letterpaper]{article}
\usepackage{emnlp2017}
\usepackage{times}
\usepackage{latexsym}
\usepackage{macro}
\usepackage{url}
\usepackage{qtree}
\usepackage{tikz-dependency}
\usepackage{etoolbox}
\patchcmd{\quote}{\rightmargin}{\leftmargin 1em \rightmargin}{}{}

\emnlpfinalcopy

\def\emnlppaperid{1219}

\title{ Structural Embedding of Syntactic Trees for Machine Comprehension }

\author{Rui Liu\thanks{Authors contributed equally to this work.}, ~~~ Junjie Hu\footnotemark[1], ~~~ Wei Wei\footnotemark[1], ~~~ Zi Yang\footnotemark[1], ~~~ Eric Nyberg \\
  School of Computer Science \\
  Carnegie Mellon University \\
  5000 Forbes Ave, Pittsburgh PA 15213, USA \\
  {\tt \{ruil, junjieh, weiwei, ziy, ehn\}@cs.cmu.edu}\\ }

\date{}

\begin{document}
\maketitle

\begin{abstract}
Deep neural networks for machine comprehension typically utilizes only word or character embeddings without explicitly taking advantage of structured linguistic information such as constituency trees and dependency trees. In this paper, we propose \textit{structural embedding of syntactic trees} (SEST), an algorithm framework to utilize structured information and encode them into vector representations that can boost the performance of algorithms for the machine comprehension. We evaluate our approach using a state-of-the-art neural attention model on the SQuAD dataset. Experimental results demonstrate that our model can accurately identify the syntactic boundaries of the sentences and extract answers that are syntactically coherent over the baseline methods.



\end{abstract}
\section{Introduction} \label{sec:intro}


Reading comprehension such as SQuAD \cite{pranav2016squad} or NewsQA \cite{trischler2016newsqa} requires identifying a span from a given context, which is an extension to the traditional question answering task, aiming at responding questions posed by human with natural language \cite{nyberg2002javelin, ferrucci2010building, liu2017thesis, yang2017thesis}. Many works have been proposed to leverage deep neural networks for such question answering tasks, most of which involve learning the query-aware context representations \cite{dhingra2016gated, seo2016bidirectional, wang2016machine, xiong2016dynamic}. 
Although deep learning based methods demonstrated great potentials for question answering, none them take syntactic information of the sentences such as \emph{constituency tree} and \emph{dependency tree} into consideration. Such techniques have been proven to be useful in many natural language understanding tasks in the past and illustrated noticeable improvements such as the work by \cite{pranav2016squad}. In this paper, we adopt similar ideas but apply them to a neural attention model for question answering. 

\begin{figure}[t]
    \centering
    \input{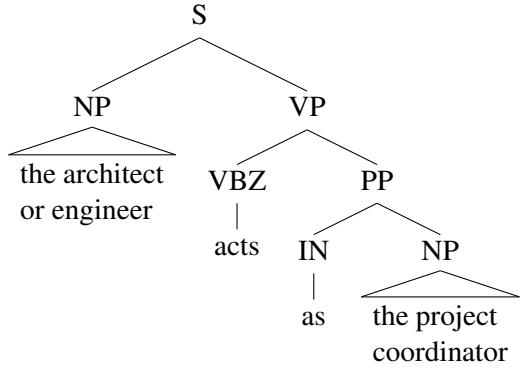}
    \caption{The constituency tree of context ``the architect or engineer acts as the project coordinator''}
    \label{fig:ctree-example}
\end{figure}

The constituency tree \cite{manning1999foundations} of a sentence defines internal nodes and terminal nodes to represent phrase structure grammars and the actual words. Figure~\ref{fig:ctree-example} illustrates the constituency tree of the sentence ``the architect or engineer acts as the project coordinator''. Here, ``the architect or engineer'' and ``the project coordinator'' are labeled as noun phrases (``NP''), which is critical for answering the question below. Here, the question asks for the name of certain occupation that can be best answered using an noun phrase. Utilizing the know ledge of a constituency relations, we can reduce the size of the candidate space and help the algorithm to identify the correct answer. 
\begin{quote}
\em
Whose role is to design the works, prepare the specifications and produce construction drawings, administer the contract, tender the works, and manage the works from inception to completion?
\end{quote}


\begin{figure*}[t]
    \centering
    \input{dtree-example}
    \caption{Partial dependency parse tree of an example context ``The Annual Conference, roughly the equivalent of a diocese in the Anglican Communion and the Roman Catholic Church or a synod in some Lutheran denominations such as the Evangelical Lutheran Church in America, is the basic unit of organization within the UMC.''}
    \label{fig:dtree-example}
\end{figure*}



On the other hand, a dependency tree \cite{manning1999foundations} is constructed based on the dependency structure of a sentence. Figure~\ref{fig:dtree-example} displays the dependency tree for sentence 
\begin{quote}
\em
The Annual Conference, roughly the equivalent of a diocese in the Anglican Communion and the Roman Catholic Church or a synod in some Lutheran denominations such as the Evangelical Lutheran Church in America, is the basic unit of organization within the UMC.
\end{quote}

``The Annual Conference'' being the subject of ``the basic unit of organization within the UMC'' provides a critical clue for the model to skip over a large chunk of the text when answering the question ``What is the basic unit of organization within the UMC''. As we show in the analysis section, adding dependency information dramatically helps identify dependency structures within the sentence, which is otherwise difficult to learn.


In this paper, we propose Structural Embedding of Syntactic Trees (SEST) that encode syntactic information structured by constituency tree and dependency tree into neural attention models for the question answering task. Experimental results on SQuAD dataset illustrates that the syntactic information helps the model to choose the answers that are both succinct and grammatically coherent, which boosted the performance on both qualitative studies and numerical results. Our focus is to show adding structural embedding can improve baseline models, rather than directly compare to published SQuAD results. Although the methods proposed in the paper are demonstrated using syntactic trees, we note that similar approaches can be used to encode other types of tree structured information such as knowledge graphs and ontology relations.

\section{Methodology}\label{sec:background}





\begin{figure}[t]
    \centering
    \includegraphics[width=0.48\textwidth]{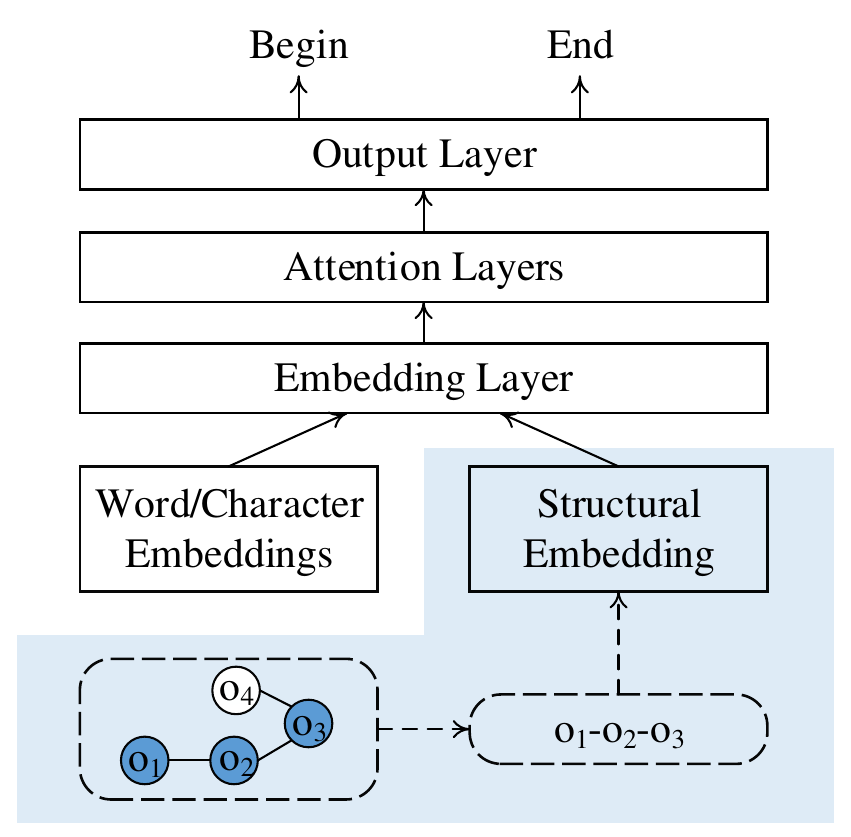}
    \caption{Model Framework. The neural network for training and testing is built by components with solid lines, which includes the embedding layer, attention layer, and output layer. The shaded area highlights the part of the framework that involves syntactic information. Components with dashed lines is an example to illustrate how syntactic information is decoded. Here a sentence is decomposed into a syntactic tree with four nodes and the syntactic information for a specific word is recorded as the path from its position in the syntactic tree to the root, i.e. $(o_1,o_2,o_3)$ in this case.
    }\label{fig:bi-att}
\end{figure}

The general framework of our model is illustrated in Figure~\ref{fig:bi-att}. Here the input of the model is the embedding of the context and question while the output is two indices \emph{begin} and \emph{end} which indicate the begin and end indices of the answer in the context space. 

The input of the model contains two parts: the word/character model and the syntactic model. The shaded portion of our model in Figure~\ref{fig:bi-att} represents the encoded syntactic information of both context and question that are fed into the model. To gain an insight of how the encoding works, consider a sentence which syntactic tree consists of four nodes $(o1,o2,o3,o4)$. A specific word is represented to be a sequence of nodes from its leave all the way to the root. We cover how this process work in detail in Section~\ref{sec:cns} and~\ref{sec:dep}. Another input that will be fed into deep learning model is the embedding information for words and characters respectively. There are many ways to convert words in a sentence into a high-dimensional embedding. We choose GloVe~\cite{pennington2014glove} to obtain a pre-trained and fixed vector for each word. Instead of using a fixed embedding, we use Convolutional Neural Networks (CNN) to model character level embedding, which values can be changed during training~\cite{kim2014convolutional}. To integrate both embeddings into the deep neural model, we feed the concatenation of them for the question and the context to be the input of the model.



The inputs are processed in the embedding layer to form more abstract representations. Here we choose a multi-layer bi-directional Long Short Term Memory (LSTM)~\cite{hochreiter1997long} to obtain more abstract representations for words in the contexts and questions. 

After that, we employ an attention layer to fuse information from both the contexts and the questions. Various matching mechanisms using attentions have been extensively studied for machine comprehension tasks ~\cite{xiong2016dynamic,seo2016bidirectional,wang2016multi,wang2016machine}. We use the Bi-directional Attention flow model ~\cite{seo2016bidirectional} which performs context-to-question and question-to-context attentions in both directions. The context-to-question attention signifies which question words are most relevant to each context word.  For each context word, the attention weight is first computed by a softmax function with question words, and the attention vector of each context word is then computed by a weighted sum of the question words' embeddings obtained from the embedding layer. The question-to-context attention summarizes a context vector by performing a soft attention with context words given the question. We then represent each context word as the concatenation of the embedding vector obtained from the embedding layer, the attention vector obtained from the context-to-question attention and the context vector obtained from the question-to-context attention. We then feed the concatenated vectors to a stacked bi-directional LSTM with two layers to obtain the final representations of context words. We note that our proposed structural embedding of syntactic trees can be easily applied to any attention approaches mentioned above. 

For the machine comprehension task in this paper, the answer to the question is a phrase in the context. In the output layer, we use two softmax functions over the output of the attention layer to predict the begin and end indices of the phrase in the context.

\section{Structural Embedding of Syntactic Tree}\label{sec:model}


We detail the procedures of two alternative implementation of our methods: the Structural Embedding of Constituency Trees model (SECT) and the Structural Embedding of Dependency Trees model (SEDT). We assume that the syntactic information has already been generated in the preprocessing step using tools such as the Stanford CoreNLP~\cite{manning-EtAl:2014:P14-5}.

\subsection{Syntactic Sequence Extraction}

We first extract a \textit{syntactic collection} $\mathcal{C}(p)$ for each word $p$, which consists of a set of nodes $\{o_{1},o_{2},\ldots,o_{d-1},o_{d}\}$ in the syntactic parse tree $\mathcal{T}$. Each node $o_i$ can be a word, a grammatical category (e.g., part-of-speech tagging), or a dependency link label, depending on the type of syntactic tree we use.
To construct syntactic embeddings, the first thing we need to do is to define a specific \textit{processing order} $\mathcal{A}$ over the syntactic collection $\mathcal{C}(p)$, in which way we can extract a \textit{syntactic sequence} $\mathcal{S}(p)$ for the word $p$.



\subsubsection{Structural Embedding of Constituency Trees (SECT)} \label{sec:cns}
The constituency tree is a syntactic parse tree constructed by phrase structure grammars~\cite{manning1999foundations}, which defines the way to hierarchically construct a sentence from words in a bottom-up manner based on constituency relations. Words in the contexts or the questions are represented by leaf nodes in the constituency tree while the non-terminal nodes are labeled by categories of the grammar. Non-terminal nodes summarize the grammatical function of the sub-tree. Figure~\ref{fig:ctree-example} shows an example of the constituency tree with ``the architect or engineer'' being annotated as a noun phrase (NP). 


A path originating from the leaf node to the root node captures the syntactic information in the constituency tree in a hierarchical way. The higher the node is, the longer span of words the sub-tree of this node covers. Hence, to extract the syntactic sequence $\mathcal{S}(p)$ for a leaf node $p$, it is reasonable to define the processing order $\mathcal{A}(p)$ from the leaf $p$ all the way to its root. For example, the syntactic sequence for the phrase ``the project coordinator" in Figure~\ref{fig:ctree-example} is detected as $\left( \text{NP, PP, VP, S}\right )$. In practice, we usually take the last hidden units of Bi-directional encoding mechanisms such as Bi-directional LSTM to represent the sequence state, as is indicated in Figure~\ref{fig:SEST} (a). We set a window size to limit the amount of information that is used in our models. For example, if we choose the window size as 2, then the syntactic sequence becomes $\left( \text{NP, PP}\right )$. This process is introduced for both performance and memory utilization consideration, which is discussed in detail in Section~\ref{sec:window_size}.   

In addition, a non-terminal node at a particular position in the syntactic sequence defines the begin and end indices of a phrase in the context. By measuring the similarity between syntactic sequences $\mathcal{S}(p)$ extracted for each word $p$ of both the question and the context, we are able to locate the boundaries of the answer span. This is done in the attention layer shown in Figure~\ref{fig:bi-att}. 




\subsubsection{Structural Embedding of Dependency Trees (SEDT)} \label{sec:dep}

The dependency tree is a syntactic tree constructed by dependency grammars~\cite{manning1999foundations}, which defines the way to connect words by directed links that represent dependencies. A dependency link is able to capture both long and short distance dependencies of words. Relations on links vary in their functions and are labeled with different categories. For example, in the dependency tree plotted in Figure~\ref{fig:dtree-example}, the link from ``unit'' to ``Conference'' indicates that the target node is a nominal subject (i.e. NSUBJ) of the source node.

The syntactic collection $\mathcal{C}(p)$ for dependency tree is defined as $p$'s children, each represented by its word embedding concatenated with a vector that uniquely identifies the dependency label.
The processing order $\mathcal{A}(p)$ for dependency tree is then defined to be the dependent's original order in the sentence. 

\begin{figure*}[t]
    \centering
    \includegraphics[width=0.9\textwidth]{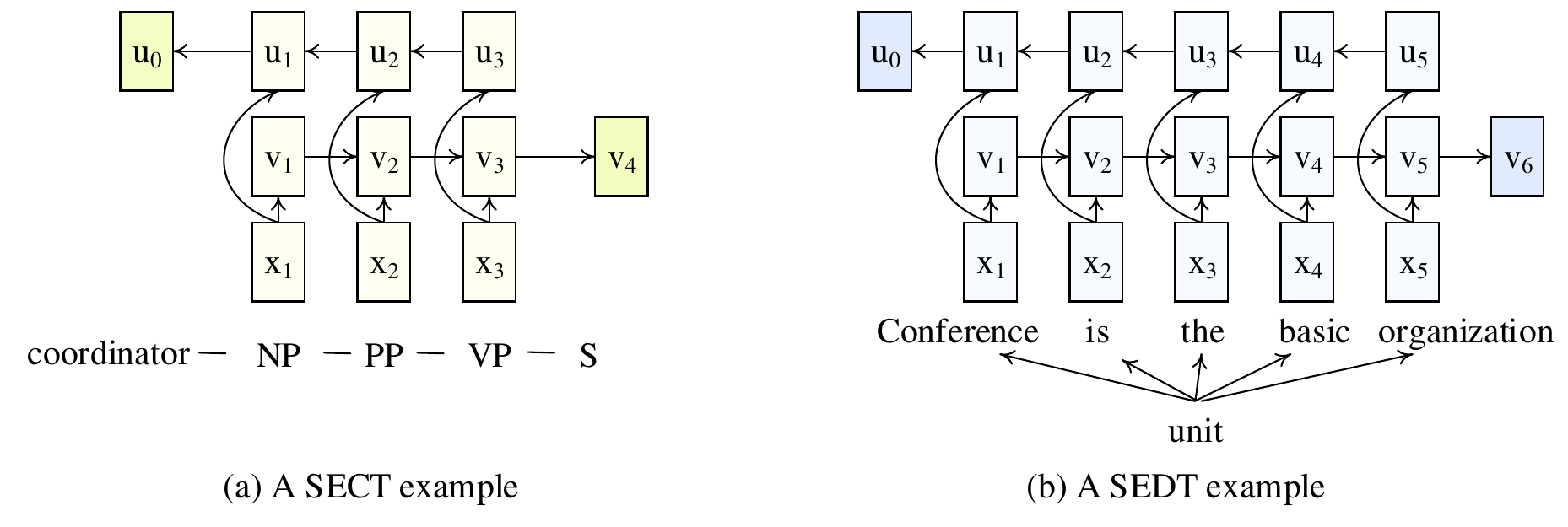}
    \caption{Two examples are used to illustrate how the syntactic information is encoded for SECT and SEDT respectively. Take Bi-directional LSTM as examples, where $\xv$ is a vector such as word embedding, $\vv$ and $\uv$ are the outputs of the forward and backward LSTMs respectively. For SECT, we encode the syntactic sequence $\left( \text{NP, PP, VP}\right )$ for the word ``coordinator" in Figure~\ref{fig:ctree-example}. We use fixed vectors for syntactic tags (e.g., NP, PP and VP), initialized with multivariate normal distribution. The final representation for the target word ``coordinator'' can be represented as the concatenation $[\Ev\wv;\textbf{u}_0;\textbf{v}_4]$, where $\Ev\wv$ is the word embedding for ``coordinator" that is 100 dimensions in our experiments and each of the encoded vector $\textbf{u}$ and $\textbf{v}$ can be 30 dimensional. For SEDT, we encode the word ``unit'' in Figure~\ref{fig:dtree-example} with its dependent nodes including ``Conference", ``is", ``the", ``basic", ``organization", ordered by their positions in the original sentence. Each word is represented with its word embedding. Similar to SECT, the final representation is the concatenation $[\Ev\wv;\textbf{u}_0;\textbf{v}_6]$, which will be sent to the input layer of a neural network.     
    }\label{fig:SEST}
\end{figure*}

Take the word ``unit'' as an example, we encode the dependency sub-tree using a Bi-directional LSTM, as indicated in Figure~\ref{fig:SEST} (b). In such as a sub-tree, since children are directly linked to the root, they are position according to the original sequence in the sentence. Similar to the syntactic encoding of C-Tree, we take the last hidden states as its embedding.

Similar to SECT, we use a window of size $l$ to limit the amount of syntactic information for the learning models by choosing only the $l$-nearest dependents, which is again reported in Section~\ref{sec:window_size}.

\subsection{Syntactic Sequence Encoding}\label{subsec:tree-path}
Similar to previous work \cite{ChoMGBBSB14, kim2014convolutional}, we use a neural network to encode a variable-length syntactic sequence into a fixed-length vector representation. The encoder can be a Recurrent Neural Network (RNN) or a Convolutional Neural Network (CNN) that learns a structural embedding for each node such that embedding of nodes under similar syntactic trees are close in their embedding space.

We can use a Bi-directional LSTM as our RNN encoder, where the hidden state $\vv_t^p$ is updated according to Eq.~\ref{eq:lstm}. Here $\xv_t^p$ is the $t^{th}$ node in the syntactic sequence of the word $p$, which is a vector that uniquely identifies each syntactic node. We obtain the structural embedding of the given word $p$, $\vv_\textnormal{Bi-LSTM}^p=\vv_T^{p}$ to be the final hidden state. 

\begin{align}
    \vv_t^p &= \textnormal{Bi-LSTM}(\vv_{t-1}^p, \xv_t^p) \label{eq:lstm}
\end{align}

Alternatively, we can also use CNN to obtain embeddings from a sequence of syntactic nodes. We denote $l$ as the length of the filter of the CNN encoder.  We define $\xv^p_{i:i+l}$ as the concatenation of the vectors from $\xv^p_i$ to $\xv^p_{i+l-1}$ within the filter. The $i^{th}$ element in the $j^{th}$ feature map can be obtained in Eq.~\ref{eq:ci}. Finally we obtain the structural embedding of the given word $p$ by $\vv_\textnormal{CNN}^p$ in Eq.~\ref{eq:u_cnn}.
\begin{align} \label{eq:ci}
\cv^{p}_{i,j}&=f(\wv_j \cdot \xv_{i:i+l-1} + b_j) \\
\vv_\textnormal{CNN}^{p}& = \textnormal{max}_{row}(\cv^p) \label{eq:u_cnn}
\end{align}
where $\wv_j$ and $b_j$ are the weight and bias of the $j^{th}$ filter respectively, $f$ is a non-linear activation function and $\textnormal{max}_{row}(\cdot)$ takes the maximum value along rows in a matrix.


\section{Experiments}

We conducted systematic experiments on the SQuAD dataset~\cite{pranav2016squad}. We compared our methods against Bi-Directional Attention Flow (BiDAF), as well as the SEST models described in Section~\ref{sec:model}. 




\subsection{Preprocessing}
A couple of preprocessing steps is performed to ensure that the deep neural models get the correct input. We segmented context and questions into sentences by using NLTK's Punkt sentence segmenter\footnote{http://www.nltk.org/api/nltk.tokenize.html}. Words in the sentences were then converted into symbols by using PTB Tokenizer\footnote{http://nlp.stanford.edu/software/tokenizer.shtml}. Syntactic information including POS tags and syntactic trees were acquired by Stanford CoreNLP utilities \cite{manning-EtAl:2014:P14-5}. For the parser, we collected constituent relations and dependency relations for each word by using tree annotation and enhanced dependencies annotation respectively. To generate syntactic sequence, we removed sequences whose first node is a punctuation (``\$'', ``:'', ``\#'', ``.'', `` '' '', `` `` '', ``,''). To use dependency labels, we removed all the subcategories (e.g., ``nmod:poss'' $\Rightarrow$ ``nmod'').

\subsection{Experiment Setting}

We run our experiments on a machine that contains a single GTX 1080 GPU with 8GB VRAM. All of the models being compared have the same settings on character embedding and word embedding. As introduced in Section~\ref{sec:background}, we use a variable character embedding with a fixed pre-trained word embedding to serve as part of the input into the model. The character embedding is implemented using CNN with a one-dimensional layer consists of 100 units with a channel size of 5. It has an input depth of 8. The max length of SQuAD is 16 which means there are a maximum 16 words in a sentence. The fixed word embedding has a dimension of 100, which is provided by the GloVe data set \cite{GloVe}. The settings for syntactic embedding are slightly different for each model. The BiDAF model does not deal with syntactic information. The POS model contains syntactic information with 39 different POS tags that serve as both input and output. For SECT and SEDT the input of the model has a size of 8 with 30 units to be output. Both of them has a maximum length size that is set to be 10 and 20 respectively, which values will be further discussed in Section~\ref{sec:window_size}. They also have two different ways to encode the syntactic information as indicated in Section~\ref{sec:model}: LSTM and CNN. We apply the same sets of parameters when we experiment them with the two models. We report the results on the SQuAD development set and the blind test set.



\subsection{Predictive Performance}
We first compared the performance of single models between the baseline approach BiDAF and the proposed SEST approaches, including SEPOS, SECT-LSTM, SECT-CNN, SEDT-LSTM, and SEDT-CNN, on the development dataset of SQuAD. For each model, we conducted 5 different single experiments and evaluated them using two metrics: ``Exact match'' (EM), which calculates the ratio of questions that are answered correctly by strict string comparison, and the F1 score, which calculates the harmonic mean of the precision and recall between predicted answers and ground true answers at the character level. As shown in Table~\ref{tab:single-model-compare}, we reported the maximum, the mean, and the standard deviation of EM and F1 scores across all single runs for each approach, and highlighted the best model using bold font. SECT-LSTM is the second best method, which confirms the predictive powers of different types of syntactic information. We could see that SEDT-LSTM model outperforms the baseline method and other proposed methods in terms of both EM and F1. 
Another observation is that our propose models achieve higher relative improvements in EM scores than F1 scores over the baseline methods, providing the evidence that syntactic information can accurately locate the boundaries of the answer. 

Moreover, we found that both SECT-LSTM and SEDT-LSTM have better performance than their CNN counterparts, which suggests that LSTM can more effectively preserve the syntactic information. As a result, we conducted further analysis of only SECT-LSTM and SEDT-LSTM models in the subsequent subsections and drop the suffix ``-LSTM'' for abbreviation. We built an ensemble model from the 5 single models for the baseline method BiDAF and our proposed methods SEPOS, SECT-LSTM, and SEDT-LSTM. The ensemble model choose the answer with the highest sum of confidence scores among the 5 single models for each question. We compared these models on both the development set and official test set and reported the results in Table~\ref{tab:ensemble-compare}. We found that the models have higher performance on the test set than the development set, which coincides with the previous results on the same data set \cite{seo2016bidirectional, xiong2016dynamic}. 

\begin{table*}[t]
 \begin{center}
 \begin{tabular}{lcccccc}
 \hline
 \multirow{3}{*}{\bf Method} & \multicolumn{4}{c}{\bf Single} & \multicolumn{2}{c}{\bf Ensemble} \\
 & \multicolumn{2}{c}{\bf EM} & \multicolumn{2}{c}{\bf F1} & \multirow{2}{*}{\bf EM} & \multirow{2}{*}{\bf F1} \\
 & \bf Max & \bf Mean ($\pm$SD) & \bf Max & \bf Mean ($\pm$SD) & & \\
 \hline
 BiDAF & 67.10 & 66.92 ($\pm$0.23)${\phantom{\bullet}}$ & 76.92 & 76.79 ($\pm$0.08) ${\phantom{\bullet}}$ & 70.97 & 79.53 \\
 SEPOS & 67.65 & 66.05 ($\pm$2.94)${\phantom{\bullet}}$ & 77.25 & 75.80 ($\pm$2.65) ${\phantom{\bullet}}$ & 71.46 & 79.70 \\
 SECT-LSTM & 67.91 & 67.65 ($\pm$0.31)${\bullet}$ & 77.47 & 77.19 ($\pm$0.21) ${\bullet}$ & 71.76 & 80.09 \\
 SEDT-LSTM & \bf 68.13 & \bf 67.89 ($\pm$0.10)${\bullet}$ & \bf 77.58 & \bf 77.42 ($\pm$0.19) $\bullet$ & \bf 72.03 &\bf 80.28 \\ 
 SECT-CNN &67.29 &64.04 ($\pm$4.28)${\phantom{\bullet}}$&76.91 &73.99 ($\pm$3.89)${\phantom{\bullet}}$ & 69.70 & 78.49\\
 SEDT-CNN &67.88 &66.53 ($\pm$1.91)${\phantom{\bullet}}$& 77.27 &76.21 ($\pm$1.67)${\phantom{\bullet}}$ & 71.58 & 79.80\\
 \hline
 \end{tabular}
 \end{center}
 \caption{\label{tab:single-model-compare}  Performance comparison on the development set. Each setting contains five runs trained consecutively. Standard deviations across five runs are shown in the parenthesis for single models. Dots indicate the level of significance.}
 \end{table*}
 
\begin{table}[t]
 \begin{center}
 \begin{tabular}{lcccc}
 \hline
 \multirow{2}{*}{\bf Method} & \multicolumn{2}{c}{\bf Single} & \multicolumn{2}{c}{\bf Ensemble} \\
 & \bf EM & \bf F1 & \bf EM & \bf F1 \\
 \hline
 BiDAF & 67.69 & 77.07 & 72.33 & 80.33 \\
 SECT-LSTM & 68.12 & 77.21 & 72.83 & 80.58 \\
 SEDT-LSTM & \bf 68.48 & \bf 77.97 & \bf 73.02 & \bf 80.84 \\ 
 \hline
 \end{tabular}
 \end{center}
 \caption{\label{tab:ensemble-compare}  Performance comparison on the official blind test set. Ensemble models are trained over the five single runs with the identical network and hyper-parameters.}
 \end{table}
 
 
 
\subsection{Contribution of Syntactic Sequence}
To take a closer look at how syntactic sequences affect the performance, we removed the character/word embedding from our model seen in Figure~\ref{fig:bi-att} and conducted experiments based on the syntactic input alone. In particular, we are interested in two aspects related to syntactic sequences: First, the ability to predict answers of questions of syntactic sequences compared to complete random sequences. Second, the amount of impacts brought by our proposed ordering introduced in Section~\ref{sec:cns} and Section~\ref{sec:dep} compared to random ordering. 






We compared the performance of the models using syntactic information along in their original order (i.e. SECT-Only and SEDT-Only) against their counterparts with the same syntactic tree nodes but with randomly shuffled order (i.e. SECT-Random-Order and SEDT-Random-Order) as well as the baselines with randomly generated tree nodes (i.e. SECT-Random and SEDT-Random). Here we choose the length of window size to be 10. The predictive results in terms of EM and F1 metrics are reported in Table~\ref{tab:tree-only}. From the table we see that both the ordering and the contents of the syntactic tree are important for the models to work properly: constituency and dependency trees achieved over 20\% boost on performance compared to the randomly generated ones and our proposed ordering also out-performed the random ordering. It also worth mentioning that the ordering of dependency trees seems to have less impact on the performance compared to that of the constituency trees. This is because sequences extracted from constituency trees contain hierarchical information, which ordering will affect the output of the model significantly. However, sequences extracted from dependency trees are all children nodes, which are often interchangeable and don't seem to be affected by ordering much. 

 \begin{table}[t]
 \begin{center}
 \begin{tabular}{l r r}\hline 
 \bf Method & \bf EM & \bf F1\\
 \hline
 SECT-Random & 5.64 & 12.85 \\
 SECT-Random-Order & 30.04 & 39.98\\
 SECT-Only & 34.21 & 44.53\\
 SEDT-Random & 0.92 & 8.82\\
 SEDT-Random-Order & 31.82 & 43.65\\
 SEDT-Only & 32.96 & 44.37\\
 \hline
 \end{tabular}
 \end{center}
 \caption{\label{tab:tree-only}  Performance comparisons of models with only syntactic information against their counterparts with randomly shuffled node sequences and randomly generated tree nodes using the SQuAD Dev set}
 \end{table}
 
\subsection{Window Size Analysis}\label{sec:window_size}
As we have mentioned in the earlier sections, limiting the window size is an important technique to prevent excessive usage on VRAM. In practice, we found that limiting the window size also benefits the performance of our models. In Table~\ref{tab:window_size-Robustness} we compared the predictive performance of SECT and SEDT models by varying the length of their window sizes from 1 to maximum on the development set. In general the results illustrate that performances of the models increase with the length of the window. However, we found that for SECT model, its mean performance reached the peak while standard deviations narrowed when window size reaches 10. We also observed that larger window size does not generate predictive results that is as good as the one with window size set to 10. This suggests that there exists an optimal window size for the constituency tree. One possible explanation is increasing the window size leads to the increase in the number of syntactic nodes in the extracted syntactic sequence. Although sub-trees might be similar between context and question, it is very unlikely that the complete trees are the same. Because of that, allowing the syntactic sequence to extend beyond the certain heights will introduce unnecessary noise into the learned representation, which will compromise the performance of the models. Similar conclusion holds for the SEDT model, which has an improved performance and decreased variance with the window size is set to 10. We did not perform experiments with window size beyond 10 for SEDT since it will consume VRAM that exceeds the capacity of our computing device. 

 
  \begin{table}[t]
 \begin{center}
 \begin{tabular}{l@{} c | c | c}\hline 
 \bf Method & \multirow{1}{*}{\bf Len} & \multicolumn{1}{c}{\bf EM} & \multicolumn{1}{|c}{\bf F1}\\\hline
  \multirow{4}{*}{SECT} & 1  &  65.58 ($\pm$ 2.58) & 75.31 ($\pm$ 2.39)\\
  & 5  &  65.74 ($\pm$ 3.77) & 75.48 ($\pm$ 3.39)\\
  & 10 & \bf 67.51 ($\pm$ 0.34) & \bf 77.14 ($\pm$ 0.39)\\
  & Max & 67.48 ($\pm$ 0.33) & 77.09 ($\pm$ 0.45) \\\hline
  \multirow{2}{*}{SEDT} & 1  &66.23 ($\pm$ 2.5) &73.85 ($\pm$ 2.22) \\
  & 10  & \bf 67.39 ($\pm$ 0.09) &\bf 76.93 ($\pm$ 0.21)\\\hline
 \end{tabular}
 \end{center}
 \caption{\label{tab:window_size-Robustness}  Performance means and standard deviations of different window sizes on the development set.}
 \end{table}
 
 

\subsection{Overlapping Analysis}
 \begin{figure}[t]
     \centering
     \usetikzlibrary{shapes,backgrounds}
\begin{tikzpicture}
  \draw (0,0) circle (1.6);
  \draw (-0.5,-0.9) circle (1.6);
  \draw (0.5,-0.9) circle (1.6);
  \node at (0,1.0) {358};
  \node at (-1.5,-1.4) {413};
  \node at (1.5,-1.4) {410};
  \node at (-1.2,0.1) {448};
  \node at (0,-1.9) {503};
  \node at (1.2,0.1) {477};
  \node at (0,-0.5) {5783};
  \node at (-2.3,1.0) {BiDAF};
  \node at (-3.0,-2.0) {SEDT};
  \node at (3.0,-2.0) {SECT};
\end{tikzpicture}
     \caption{Venn Diagram on the number of correct answers predicted by BiDAF, SECT and SEDT}
     \label{fig:venn3}
 \end{figure}
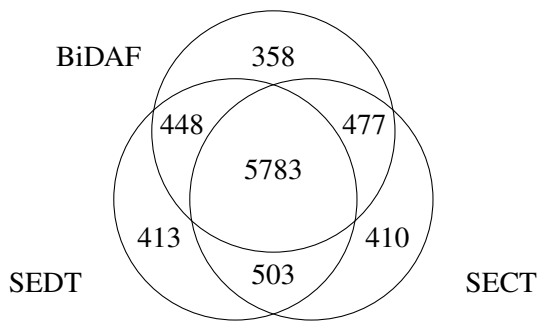

\begin{table*}[t]
    \centering
    \begin{tabular}{p{0.4\textwidth}p{0.30\textwidth}p{0.19\textwidth}}
    \bf Question & \bf BiDAF & \bf SECT \\
    \hline
    Whose role is to design the works, prepare the specifications and produce construction drawings, administer the contract, tender the works, and manage the works from inception to completion? & {[}the architect or engineer{]$_\texttt{NP}$} {[}acts as {[}the project coordinator{]$_\texttt{NP}$}{]$_\texttt{VP}$} & {[}the architect or engineer{]$_\texttt{NP}$} \\
    \hline
    What did Luther think the study of law meant? & {[}represented {[}uncertainty{]$_\texttt{NP}$}{]$_\texttt{VP}$} & {[}uncertainty{]$_\texttt{NP}$} \\
    \hline
    What caused the dynamos to be burnt out? & {[}the powerful high frequency currents{]$_\texttt{NP}$} {[}set up {[}in {[}them{]$_\texttt{NP}$}{]$_\texttt{PP}$}{]$_\texttt{VP}$} & {[}powerful high frequency currents{]$_\texttt{NP}$}\\
    \hline
    \end{tabular}
    \caption{Questions that are correctly answered by SECT but not BiDAF}
    \label{tab:bidaf-ctree-output}
\end{table*}

\begin{table*}[t]
    \centering
    \begin{tabular}{p{0.19\textwidth}p{0.44\textwidth}p{0.15\textwidth}p{0.09\textwidth}}
    \bf Question & \bf Context & \bf BiDAF & \bf SEDT \\
    \hline
    In the layered model of the Earth, the mantle has two layers below it. What are they? & These advances led to the development of a layered model of the Earth, with a crust and lithosphere on top, the mantle below (separated within itself by seismic discontinuities at 410 and 660 kilometers), and the outer core and inner core below that. & seismic discontinuities at 410 and 660 kilometers), and the outer core and inner core & the outer core and inner core \\
    \hline
    What percentage of farmland grows wheat? & More than 50\% of this area is sown for wheat, 33\% for barley and 7\% for oats. & 33\% & 50\% \\
    \hline
    What is the basic unit of organization within the UMC? & The Annual Conference, roughly the equivalent of a diocese in the Anglican Communion and the Roman Catholic Church or a synod in some Lutheran denominations such as the Evangelical Lutheran Church in America, is the basic unit of organization within the UMC. & Evangelical Lutheran Church in America & The Annual Conference \\
    \hline
    \end{tabular}
    \caption{Questions that are correctly answered by SEDT but not BiDAF}
    \label{tab:bidaf-dtree-output}
\end{table*}

To further understand the performance benefits of incorporating syntactic information into the question answering problem, we can take a look at the questions on which models disagree. Figure~\ref{fig:venn3} is the Venn Diagram on the questions that have been corrected identified by SECT, SEDT and the baseline BiDAF model. Here we see that the vast majority of the correctly answered questions are shared across all three models. The rest of them indicates questions that models disagree and are distributed fairly evenly.

To understand the types of the questions that syntactic models can do better, we extracted three questions that were correctly answered by SECT and SEDT but not the baseline model. In Table~\ref{tab:bidaf-ctree-output}, all of the three questions are ``Wh-questions'' and expect the answer of a noun phrase (NP). Without knowing the syntactic information, BiDAF answered questions with unnecessary structures such as verb phrases (vp) (e.g. ``acts as $\cdots$ '', ``represented $\cdots$ '') or prepositional phrases (pp) (e.g. ``in $\cdots$ '') in addition to NPs (e.g. ``the architect engineer'', ``uncertainty'' and ``powerful high frequency currents'') that normal human would answer. For that reason, answers provided by BiDAF failed the exact match although its answers are semantically equivalent to the ones provided by SECT. Having incorporated constituency information provided an huge advantage in inferring the answers that are most natural for a human. 


The advantages of using the dependency tree in our model can be illustrated using the questions in Table~\ref{tab:bidaf-dtree-output}. Here again we listed the ones that are correctly identified by SEDT but not BiDAF. As we can see that the answer provided by BiDAF for first question broke the parenthesis incorrectly, this problem that can be easily solved by utilizing dependency information. In the second example, BiDAF failed to identify the dependency structures between ``50\%'' and the keyword being asked ``wheat'', which resulted in an incorrect answer that has nothing to do with the question. SEDT, on the other hand, answered the question correctly. In the third question, the key to the answer is to correctly identify the subject of question phrase ``is the basic unit of organization''. Using the dependency tree as illustrated in Figure~\ref{fig:dtree-example}, SEDT is able to identify the subject phrase correctly, namely ``The Annual Conference''. However, BiDAF failed to anwer the question correctly and selected a noun phrase as the answer.

\section{Related Work}\label{sec:related}

\textbf{Reading Comprehension.}
Reading comprehension is a challenging task in NLP research. Since the release of the SQuAD data set, many works have been done to construct models on this massive question answering data set. Rajpurkar et. al. are among the first authors to explore the SQuAD. They used logistic regression with pos tagging information~\cite{pranav2016squad} and provided a strong baseline for all subsequent models. A steep improvement was given by the RaSoR model \cite{lee2016learning} which utilized recurrent neural networks to consider all possible subphrases of the context and evaluated them one by one. To avoid comparing all possible candidates and to improve the performance, Match-LSTM ~\cite{wang2016machine} was proposed by using a pointer network~\cite{vinyals2015pointer} to extract the answer span from the context. The same idea was taken to the BiDAF~\cite{seo2016bidirectional} model by introducing a bi-directional attention mechanism. Despite the above-mentioned strong models for the machine comprehension task, none of them considers syntactic information into their prediction models.



\textbf{Representations of Texts and Words.} 
One of the main issues in reading comprehension is to identify the latent representations of texts and words~\cite{cui2016attention, lee2016learning, wang2016multi, xiong2016dynamic, yu2016chunk}. Many pre-trained libraries such as word2vec \cite{mikolov2013distributed} and Glove \cite{GloVe} have been widely used to map words into a high dimensional embedding space. Another approach is to generate embeddings by using neural networks models such as Character Embedding \cite{kim2014convolutional} and Tree-LSTM~\cite{Tai15improvedsemantic}. One thing that worth mentioning is that although Tree-LSTM does utilize syntactic information, it targets at the phrases or sentences level embedding other than the word level embedding we have discussed in this paper. Many machine comprehension models include both pre-trained embeddings and variable embeddings that can be changed through a training stage \cite{seo2016bidirectional}.


\section{Conclusion}
In this paper, we proposed methods to embed syntactic information into the deep neural models to improve the accuracy of our model in the machine comprehension task. We formally defined our SEST framework and proposed two instances to it: the structural embedding of constituency trees (SECT) and the structural embedding of dependency trees (SEDT). Experimental results on SQuAD data set showed that our proposed approaches outperform the state-of-the-art BiDAF model, proving that the proposed embeddings play a significant part in correctly identifying answers for the machine comprehension task. In particular, we found that our model can perform especially well on exact match metrics, which requires syntactic information to accurately locate the boundaries of the answers. 
Similar approaches can be used to encode other tree structures such as knowledge graphs and ontology relations. 

This work opened several potential new lines of research: 1) In the experiments of our paper we utilized the BiDAF model to retrieve answers from the context. Since there are no structures in the BiDAF models to specifically optimize for syntactic information, an attention mechanism that is designed for to utilize syntactic information should be studied. 2) Another direction of research is to incorporate SEST with deeper neural networks such as VD-CNN \cite{conneau2017very} to improve learning capacity for syntactic embedding. 3) Tree structured information such as knowledge graphs and ontology structure should be studied and improve question answering tasks using similar techniques to the ones proposed in the paper.  

\newpage

\bibliography{ref}

\begin{thebibliography}{}
\expandafter\ifx\csname natexlab\endcsname\relax\def\natexlab#1{#1}\fi

\bibitem[{Cho et~al.(2014)Cho, van Merrienboer, G{\"{u}}l{\c{c}}ehre, Bahdanau,
  Bougares, Schwenk, and Bengio}]{ChoMGBBSB14}
Kyunghyun Cho, Bart van Merrienboer, {\c{C}}aglar G{\"{u}}l{\c{c}}ehre, Dzmitry
  Bahdanau, Fethi Bougares, Holger Schwenk, and Yoshua Bengio. 2014.
\newblock Learning phrase representations using {RNN} encoder-decoder for
  statistical machine translation.
\newblock In {\em EMNLP\/}. pages 1724--1734.

\bibitem[{Conneau et~al.(2017)Conneau, Schwenk, Barrault, and
  Lecun}]{conneau2017very}
Alexis Conneau, Holger Schwenk, Lo{\"\i}c Barrault, and Yann Lecun. 2017.
\newblock Very deep convolutional networks for text classification.
\newblock In {\em EACL\/}.

\bibitem[{Cui et~al.(2016)Cui, Chen, Wei, Wang, Liu, and Hu}]{cui2016attention}
Yiming Cui, Zhipeng Chen, Si~Wei, Shijin Wang, Ting Liu, and Guoping Hu. 2016.
\newblock Attention-over-attention neural networks for reading comprehension.
\newblock {\em arXiv preprint arXiv:1607.04423\/} .

\bibitem[{Dhingra et~al.(2016)Dhingra, Liu, Cohen, and
  Salakhutdinov}]{dhingra2016gated}
Bhuwan Dhingra, Hanxiao Liu, William~W Cohen, and Ruslan Salakhutdinov. 2016.
\newblock Gated-attention readers for text comprehension.
\newblock {\em arXiv preprint arXiv:1606.01549\/} .

\bibitem[{Ferrucci et~al.(2010)Ferrucci, Brown, Chu-Carroll, Fan, Gondek,
  Kalyanpur, Lally, Murdock, Nyberg, Prager et~al.}]{ferrucci2010building}
David Ferrucci, Eric Brown, Jennifer Chu-Carroll, James Fan, David Gondek,
  Aditya~A Kalyanpur, Adam Lally, J~William Murdock, Eric Nyberg, John Prager,
  et~al. 2010.
\newblock Building watson: An overview of the deepqa project.
\newblock {\em AI magazine\/} 31(3):59--79.

\bibitem[{Hochreiter and Schmidhuber(1997)}]{hochreiter1997long}
Sepp Hochreiter and J{\"u}rgen Schmidhuber. 1997.
\newblock Long short-term memory.
\newblock {\em Neural computation\/} .

\bibitem[{Kim(2014)}]{kim2014convolutional}
Yoon Kim. 2014.
\newblock Convolutional neural networks for sentence classification.
\newblock In {\em EMNLP\/}. pages 1746--1751.

\bibitem[{Lee et~al.(2016)Lee, Kwiatkowski, Parikh, and Das}]{lee2016learning}
Kenton Lee, Tom Kwiatkowski, Ankur Parikh, and Dipanjan Das. 2016.
\newblock Learning recurrent span representations for extractive question
  answering.
\newblock {\em arXiv preprint arXiv:1611.01436\/} .

\bibitem[{Liu(2017)}]{liu2017thesis}
Rui Liu. 2017.
\newblock {\em A Phased Ranking Model for Information Systems\/}.
\newblock Ph.D. thesis, Carnegie Mellon University.

\bibitem[{Manning et~al.(1999)Manning, Sch{\"u}tze
  et~al.}]{manning1999foundations}
Christopher~D Manning, Hinrich Sch{\"u}tze, et~al. 1999.
\newblock {\em Foundations of statistical natural language processing\/},
  volume 999.
\newblock MIT Press.

\bibitem[{Manning et~al.(2014)Manning, Surdeanu, Bauer, Finkel, Bethard, and
  McClosky}]{manning-EtAl:2014:P14-5}
Christopher~D. Manning, Mihai Surdeanu, John Bauer, Jenny Finkel, Steven~J.
  Bethard, and David McClosky. 2014.
\newblock \href{http://www.aclweb.org/anthology/P/P14/P14-5010}{The {Stanford}
  {CoreNLP} natural language processing toolkit}.
\newblock In {\em Association for Computational Linguistics (ACL) System
  Demonstrations\/}. pages 55--60.
\newblock
  \href{http://www.aclweb.org/anthology/P/P14/P14-5010}{http://www.aclweb.org/anthology/P/P14/P14-5010}.

\bibitem[{Mikolov et~al.(2013)Mikolov, Sutskever, Chen, Corrado, and
  Dean}]{mikolov2013distributed}
Tomas Mikolov, Ilya Sutskever, Kai Chen, Greg~S Corrado, and Jeff Dean. 2013.
\newblock Distributed representations of words and phrases and their
  compositionality.
\newblock In {\em Advances in neural information processing systems\/}. pages
  3111--3119.

\bibitem[{Nyberg et~al.(2002)Nyberg, Mitamura, Carbonell, Callan,
  Collins-Thompson, Czuba, Duggan, Hiyakumoto, Hu, Huang
  et~al.}]{nyberg2002javelin}
Eric Nyberg, Teruko Mitamura, Jaime~G Carbonell, Jamie Callan, Kevyn
  Collins-Thompson, Krzysztof Czuba, Michael Duggan, Laurie Hiyakumoto, N~Hu,
  Yifen Huang, et~al. 2002.
\newblock The javelin question-answering system at trec 2002.
\newblock In {\em TREC\/}.

\bibitem[{Pennington et~al.(2014{\natexlab{a}})Pennington, Socher, and
  Manning}]{GloVe}
Jeffrey Pennington, Richard Socher, and Christopher Manning.
  2014{\natexlab{a}}.
\newblock Glove: Global vectors for word representation.
\newblock In {\em EMNLP\/}. Association for Computational Linguistics, pages
  1532--1543.

\bibitem[{Pennington et~al.(2014{\natexlab{b}})Pennington, Socher, and
  Manning}]{pennington2014glove}
Jeffrey Pennington, Richard Socher, and Christopher~D. Manning.
  2014{\natexlab{b}}.
\newblock \href{http://www.aclweb.org/anthology/D14-1162}{Glove: Global vectors
  for word representation}.
\newblock In {\em Empirical Methods in Natural Language Processing (EMNLP)\/}.
  pages 1532--1543.
\newblock
  \href{http://www.aclweb.org/anthology/D14-1162}{http://www.aclweb.org/anthology/D14-1162}.

\bibitem[{Rajpurkar et~al.(2016)Rajpurkar, Zhang, Lopyrev, and
  Liang}]{pranav2016squad}
Pranav Rajpurkar, Jian Zhang, Konstantin Lopyrev, and Percy Liang. 2016.
\newblock Squad: 100,000+ questions for machine comprehension of text.
\newblock In {\em EMNLP\/}.

\bibitem[{Seo et~al.(2017)Seo, Kembhavi, Farhadi, and
  Hajishirzi}]{seo2016bidirectional}
Minjoon Seo, Aniruddha Kembhavi, Ali Farhadi, and Hannaneh Hajishirzi. 2017.
\newblock Bidirectional attention flow for machine comprehension.
\newblock In {\em ICLR\/}.

\bibitem[{Tai et~al.(2015)Tai, Socher, and Manning}]{Tai15improvedsemantic}
Kai~Sheng Tai, Richard Socher, and Christopher~D. Manning. 2015.
\newblock Improved semantic representations from tree-structured long
  short-term memory networks.
\newblock In {\em ACL\/}.

\bibitem[{Trischler et~al.(2016)Trischler, Wang, Yuan, Harris, Sordoni,
  Bachman, and Suleman}]{trischler2016newsqa}
Adam Trischler, Tong Wang, Xingdi Yuan, Justin Harris, Alessandro Sordoni,
  Philip Bachman, and Kaheer Suleman. 2016.
\newblock Newsqa: A machine comprehension dataset.
\newblock {\em arXiv preprint arXiv:1611.09830\/} .

\bibitem[{Vinyals et~al.(2015)Vinyals, Fortunato, and
  Jaitly}]{vinyals2015pointer}
Oriol Vinyals, Meire Fortunato, and Navdeep Jaitly. 2015.
\newblock Pointer networks.
\newblock In {\em NIPS\/}. pages 2692--2700.

\bibitem[{Wang and Jiang(2016)}]{wang2016machine}
Shuohang Wang and Jing Jiang. 2016.
\newblock Machine comprehension using match-lstm and answer pointer.
\newblock {\em arXiv preprint arXiv:1608.07905\/} .

\bibitem[{Wang et~al.(2016)Wang, Mi, Hamza, and Florian}]{wang2016multi}
Zhiguo Wang, Haitao Mi, Wael Hamza, and Radu Florian. 2016.
\newblock Multi-perspective context matching for machine comprehension.
\newblock {\em arXiv preprint arXiv:1612.04211\/} .

\bibitem[{Xiong et~al.(2017)Xiong, Zhong, and Socher}]{xiong2016dynamic}
Caiming Xiong, Victor Zhong, and Richard Socher. 2017.
\newblock Dynamic coattention networks for question answering.
\newblock In {\em ICLR\/}.

\bibitem[{Yang(2017)}]{yang2017thesis}
Zi~Yang. 2017.
\newblock {\em Analytics Meta Learning\/}.
\newblock Ph.D. thesis, Carnegie Mellon University.

\bibitem[{Yu et~al.(2016)Yu, Zhang, Hasan, Yu, Xiang, and Zhou}]{yu2016chunk}
Yang Yu, Wei Zhang, Kazi Hasan, Mo~Yu, Bing Xiang, and Bowen Zhou. 2016.
\newblock End-to-end answer chunk extraction and ranking for reading
  comprehension.
\newblock {\em arXiv preprint arXiv:1610.09996\/} .

\end{thebibliography}
\bibliographystyle{emnlp_natbib}
\end{document}


\maketitle

We detail the neural network architecture that we used in Section 4, focusing on describing the attention layer and the output layer.

\section{Attention Layer}
Notice that our proposed structural embedding of syntactic trees can be easily applied to any attention approaches~\cite{xiong2016dynamic,seo2016bidirectional,wang2016multi,wang2016machine}. In this paper, we use the Bi-directional Attention flow~\cite{seo2016bidirectional} which performs the context-to-question and question-to-context attentions in both directions.

The inputs to the attention layer are the embeddings of context words and question words obtained from the embedding layer.  We denote the embeddings of $T$ context words as $\Hv=[\hv_1; \cdots; \hv_T]\in\Rb^{d\times T}$ where $\hv_t\in\Rb^d~\forall t$, and denote the embeddings of $J$ question words as $\Uv=[\uv_1; \cdots; \uv_J]\in\Rb^{d\times J}$ where $\uv_j\in\Rb^d~\forall j$. We first calculate the similarity matrix $\Sv\in\Rb^{T\times J}$ where $S_{tj}$ measures the similarity between the $t$-th context word and the $j$-th question word by Equation~\ref{eq:sim}. 

\begin{align}\label{eq:sim}
    S_{tj} = \Wv_{\Sv}^T[\hv_t; \uv_j; \hv_t \circ \uv_j]
\end{align}
where $\Wv_{\Sv}\in\Rb^{3d}$ is a trainable parameter, $\circ$ is the elementwise multiplication and $[;]$ is the vector concatenation along rows.

The context-to-question attention performs a soft attention over all question words for each context word. The attention weight is computed by $\av_t=\textnormal{softmax}(\Sv_{t:})\in\Rb^J$ where $\Sv_{t:}$ is the $t$-th row of $\Sv$. Then each attended vector for each context word is calculated by $\tilde{\hv}_t=\sum_{j=1}^J a_{tj} \uv_j$.

The question-to-context attention first calculates the attention weight over the context words by $\bv = \textnormal{softmax}(\textnormal{max}_{col}(\Sv)) \in \Rb^T$, where max$_{col}$ returns the maximum values across the columns in a matrix. Then the summarized context vector can be calculated by $\hat{\hv}=\sum_{t=1}^T b_t \hv_t\in \Rb^{d}$.  

Finally, we combine three input vectors, i.e., $\hv_t$, $\tilde{\hv}_t$ and $\hat{\hv}$, for the $t$-th context word by $\gv_t = [\hv_t; \tilde{\hv}_t; \hv_t \circ \tilde{\hv}_t; \hv_t \circ \hat{\hv}]\in \Rb^{4d}$. The question-aware representation of the context, i.e., $\Gv=[\gv_1, \cdots, \gv_T]\in\Rb^{4d\times T}$, is further encoded by a stacked Bi-directional LSTM with two layers to obtained $\Mv=[\mv_1; \cdots, \mv_T]in \Rb^{d\times T}$.

\section{Output Layer}
For the machine comprehension task, the begin and end indices over the entire context is predicted by Equation~\ref{eq:start} and~\ref{eq:end} respectively. 
\begin{align}\label{eq:start}
    \pv^1 = \textnormal{softmax}(\Wv^T_{\pv^1} [G;M]) \\\label{eq:end}
    \pv^2 = \textnormal{softmax}(\Wv^T_{\pv^2} [G;M^2])
\end{align}
where $\Wv_{\pv^1}$ and $\Wv_{\pv^2}$ are trainable parameters and $\Mv^2$ is obtained by passing $\Mv$ to another bidirectional LSTM.

\section{Training}
We 
\bibliography{ref}
\bibliographystyle{emnlp_natbib}